\documentclass[letterpaper, 10 pt, conference]{ieeeconf}  % Comment this line out if you need a4paper
\usepackage{graphicx}
\usepackage{float}

\IEEEoverridecommandlockouts                              % This command is only needed if 
\title{\LARGE \bf
Human Emotion-Mediated Soft Robotic Arts: Exploring the Intersection of Human Emotions, Soft Robotics and Arts 
}

\author{Saitarun Nadipineni$^{1}$, Chenhao Hong$^{1}$, Tanishtha Ramlall$^{1}$, Chapa Sirithunge$^{2}$, Kaspar Althoefer$^{1}$, Fumiya Iida$^{2}$ \\ and  Thilina Dulantha Lalitharatne$^{1}$ 
\thanks{$^{1}$ School of Engineering and Materials Science in Queen Mary University of London, United Kingdom.
        {\tt\small c.hong@se23.qmul.ac.uk, t.ramlall@se21.qmul.ac.uk,\{s.nadipineni,\newline k.althoefer,t.lalitharatne\}@qmul.ac.uk}}%
\thanks{$^{2}$Department of Engineering, University of Cambridge,
        Cambridge CB2 1PZ, United Kingdom.
        {\tt\small \{csh66, fi224\}@cam.ac.uk }}
        \thanks{This work was supported by 101034337 — FUTUREROADS — H2020-MSCA-COFUND-2020.}
}

\begin{document}

\maketitle
\thispagestyle{empty}
\pagestyle{empty}
%%%%%%%%%%%%%%%%%%%%%%%%%%%%%%%%REMARKS
% 1. To complete embodiment, better to integrate the color changing platform as in,
% flower >> user's stress level
% color changing platform > environment stimuli

%I suggest adding some stamen which has color changing ability in the centre of the flower
%%%%%%%%%%%%%%%%%%%%%%%%%%%%%%%%%%%%%%%%%%%%%%%%%%%%%%%%%%%%%%%%%%%%%%%%%%%%%%%%

\begin{abstract}

%Soft robotics has emerged as a versatile field with applications across a wide range of domains, from healthcare to industrial automation, and more recently, art and interactive installations. The inherent flexibility, adaptability, and safety of soft robots make them ideal for applications that require delicate, organic, and lifelike movement, allowing for immersive and responsive interactions. This study explores intersection of human emotions, soft robotics and arts to establish and create a new forms of human emotion mediated soft robotic arts. In this paper, we introduce two soft embodiments; a soft character and a soft flower as an art display that dynamically responds to brain signals based on alpha waves, reflecting different emotion levels. We present how the human emotions can be measured as alpha waves based on brain/EEG signals, how we map the alpha waves with the dynamic movements of the two soft embodiment and demonstrate our proposed concept using experiments. This work shows how soft robotics can embody human emotional states, offering a new medium for artistic expression and interaction and can be used as a demonstration of how art displays can be embodied, multipurpose and a representation of both life and the environment. The findings highlight the potential for bio-inspired design and its implications in human-robot interactions, emotional robotics, and the fusion of technology and art.

Soft robotics has emerged as a versatile field with applications across various domains, from healthcare to industrial automation, and more recently, art and interactive installations. The inherent flexibility, adaptability, and safety of soft robots make them ideal for applications that require delicate, organic, and lifelike movement, allowing for immersive and responsive interactions. This study explores the intersection of human emotions, soft robotics, and art to establish and create new forms of human emotion-mediated soft robotic art. In this paper, we introduce two soft embodiments: a soft character and a soft flower as an art display that dynamically responds to brain signals based on alpha waves, reflecting different emotion levels. We present how human emotions can be measured as alpha waves based on brain/EEG signals, how we map the alpha waves to the dynamic movements of the two soft embodiments, and demonstrate our proposed concept using experiments. The findings of this study highlight how soft robotics can embody human emotional states, offering a new medium for insightful artistic expression and interaction, and demonstrating how art displays can be embodied.

\end{abstract}

%%%%%%%%%%%%%%%%%%%%%%%%%%%%%%%%%%%%%%%%%%%%%%%%%%%%%%%%%%%%%%%%%%%%%%%%%%%%%%%%
\section{INTRODUCTION}

% Soft robotics offers several key features that make it ideal for artistic applications. Its flexibility and fluid movement allow for lifelike, organic motions, while its tactile aesthetic encourages touch and interaction. The robots’ ability to adapt to organic shapes makes them perfect for dynamic or evolving installations. Their inherent safety allows for close human interaction without risk, making them suitable for public spaces. Additionally, soft robots can easily be integrated with other media like sensors or lighting systems, creating interactive experiences that respond to environmental or emotional stimuli. Finally, soft robots can mimic biological behaviors, offering a powerful way to express the essence of organic life in art.

% Recent research has highlighted the growing potential of robots to engage in interactive tasks where humans tend to trust them more, particularly in specific applications, as opposed to tasks like escorting children to school, which still elicit greater reservations \cite{sirithunge2023we}. 

%  Outsourcing functionality of soft robots directly to the body morphology could bring its control closer to biological systems \cite{hauser2023leveraging}.

%  Flora deployed in this work implement its intelligence in the form of adaptive functionality and  mechanics \cite{kortman2024perspectives}.

 Emotions are a fundamental aspect of the human experience, influencing everything from interpersonal relationships to decision-making. They are deeply intertwined with our perception of the world, helping us interpret social cues, empathize with others, and navigate different social environments. Therefore, the ability to express emotions positively contributes to personal well-being and is the foundation of human connection and understanding. Yet for individuals with speech impairments, such as those with Dysarthria \cite{alhinti2021acoustic} and articulation disorders, verbal emotional expression can be challenging. This is where Electroencephalography (EEG) technology offers exciting possibilities to help understand and express emotions differently, tailored to individual needs. 

 \begin{figure}[htbp] 
    \centerline{
    \includegraphics[scale=0.29]{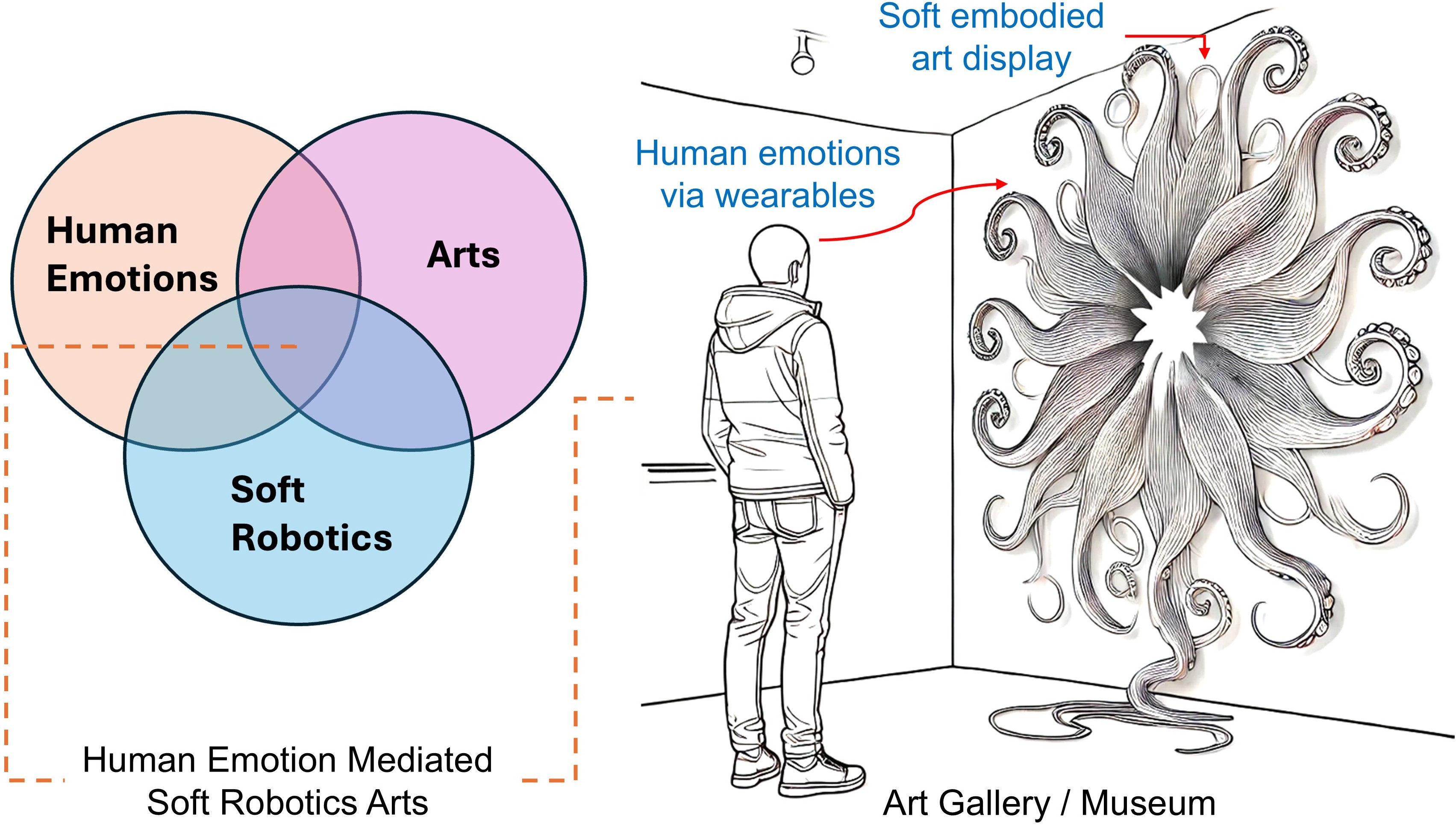}}
    \caption{Overview of the proposed concept: Human emotions through brain signals, the fusion of human emotions with art, and soft robotics with art have been reported; however, the intersection of these three fields has been largely under-explored. In this paper, we propose a concept of human emotion-mediated soft robotic art. We exploit advancements in human emotion detection via wearables, such as brain-computer interfaces, to enable dynamic and immersive art displays in galleries or museums through soft embodied art displays. As opposed to rigid counterparts, soft embodiments offer flexibility, fluid, lifelike movement, organic motions, and safety, allowing for close interactions. \textit{[The initial image of the art gallery/museum was generated with ChatGPT 4.0 (DALL-E) using the prompt: 'Draw an outline picture of a person looking at a wall with soft robotics art moving'.]}}
    \label{sys_overview_fig}
    \vspace*{-3mm}
\end{figure}

 EEG allows for the recording and analysis of brainwave patterns associated with emotional states. Hence why EEG-based emotional analysis has gained traction in various fields, such as emotional health care and human-computer interaction \cite{hamzah2024eeg}. In recent years, more studies have explored the relationship between EEG signals and emotion recognition and have made significant progress, especially in the recognition of complex emotions such as happiness, sadness, or anxiety \cite{alarcao2017emotions, suhaimi2020eeg, 10498398}. In therapy, real-time EEG feedback can help therapists monitor patients' emotional responses, guiding interventions and offering tailored support \cite{Tacca2024treatment}. 

 EEG has been used to provide emotional expression through art \cite{artfund2023, neuroelectrics2024}. EEG and art represent a fascinating intersection where neuroscience meets creative expression, allowing brainwaves to influence and shape artistic experiences in real-time. The art performance, Eunoia, produced by Lisa Park is a perfect example of EEG being used for artistic purposes \cite{Lee_2022}. This performance used a commercial EEG sensor to manifest the artist's mental state into sounds to manipulate water \cite{Jobson2013water}. These examples illustrate how EEG in arts can provide a non-verbal outlet for human emotion.

 EEG in soft robotics is an emerging field that combines brain-computer interface (BCI) technology with flexible robots \cite{yang2020flexible} to create responsive, interactive systems. Soft robots can adapt to various unstructured environments and interact with humans safely \cite{yasa2023overview}. BCI is a system that transfers brain signals into output commands for an external device \cite{neurosciences2020}, bypassing traditional neuromuscular or peripheral nerve pathways \cite{shih2012medicine}. Combined with soft robotics, they can be programmed to respond to a person's mental and emotional state. One example of leveraging EEG's real-time responsiveness was the development of a multimodal human-machine interface (MHMI) to control a robotic hand \cite{zhang2019hand}. Integrating EEG and soft robotics allows for real-time responses for the dynamic control of lifelike movements to bridge the gap between human intent and robotic action.
 
 Soft robotics for aesthetics and arts has gained attention in recent years \cite{Jørgensen1}. Soft robotics offers several key features that make it ideal for artistic applications. Its flexibility and fluid movement \cite{yang2020flexible} allow for lifelike, organic motions, while its tactile aesthetic encourages touch and interaction. The robots’ ability to adapt to organic shapes makes them perfect for dynamic or evolving installations \cite{wang2024advancements}. Their inherent safety allows for close human interaction without risk, making them suitable for public spaces \cite{lee2017soft}. Additionally, soft robots can easily be integrated with other media like sensors or lighting systems, creating interactive experiences that respond to environmental or emotional stimuli. Finally, soft robots can mimic biological behaviors, offering a powerful way to express the essence of organic life in art \cite{ren2021biology} \cite{yang2022bioinspired}.

 The intersection of human emotion, soft robotics, and the arts represents a largely underexplored field of study. Integrating these domains offers a unique opportunity to create systems that, not only respond to human emotional states but also embody and express these states. An overview of the proposed concept is shown in Fig. 1.
 
 In this paper, we have attempted to introduce EEG-based emotional feedback for the control of different soft embodiments, which in turn can act as an expressive medium. We introduce two representations of soft embodiments, a soft flower in terms of an art display and a soft character that dynamically responds to emotional changes of the human user measured based on the brain signals. 
 %In the case of Flora combines sensory feedback from BCIs with the adaptive mechanics of soft robotics, embodying naturalistic movements and responses reminiscent of real flora \cite{kortman2024perspectives}. 
 We used two soft embodiments to showcase the potential of using brain-mediated soft robotic arts through multiple mediums. Through this integration, both the soft flower and the soft character exemplify the potential of soft robotics to bridge technology and artistic expression, creating an innovative, responsive platform where human emotions, art, and robotics converge. By translating real-time brain signals into the adaptive behaviors of a soft robot, this work opens a new frontier in nonverbal, emotion-based communication, offering a platform that uniquely blends human experience, technology, and art. To our knowledge, we are the first group to explore human emotion-mediated soft robotic arts, as we found no existing literature in this area.

The rest of the paper is structured as follows. In the methods section, we present details on measuring human emotions using brain alpha waves, the development of the two soft embodiments—the soft character and the soft flower—and mapping the alpha waves to the dynamic movements of the two soft embodiments. Section III presents the experiments and results, along with a discussion of the proposed concept and the limitations of the current study. Finally, the paper concludes with potential future directions in Section IV.

\section{METHODS}

\subsection{Human emotions through brain alpha waves}\label{eeg_fft_sec}

Alpha waves are a type of low-frequency brainwave, typically ranging from 8 to 13 Hz, and are closely associated with a relaxed state of mind \cite{bazanova2014interpreting}. Alpha waves commonly appear during moments of rest with closed eyes or during meditation, and their intensity and frequency fluctuations can reflect a person’s level of relaxation or anxiety. This makes alpha waves widely used in emotion recognition \cite{ismail2016human}. Research has shown a clear relationship between changes in alpha waves and emotional states, particularly in reducing tension, relaxing the mind, and managing stress \cite{iwaki2012eeg, nordin2022classification}. This characteristic makes alpha waves an important signal for studying emotional responses in neuroscience and BCI fields. Many emotion recognition systems and interactive art projects measure alpha waves to gain direct feedback on the brain’s emotional state. They are particularly prominent when a person’s eyes are closed, as visual and external stimuli are minimized, allowing the brain to settle into a more relaxed rhythm. When the eyes are open, alpha wave activity typically decreases in a process known as alpha-blocking, where the brain shifts focus to process external visual information, disrupting the calm state associated with alpha waves \cite{kaiser2005basic}.

Alpha waves are most detectable in the occipital cortex, located in the rear of the brain, where visual processing primarily occurs \cite{halgren2019generation}. This region is especially sensitive to visual input, so alpha activity is strongest when visual stimulation is reduced, such as when the eyes are closed. Because of this, electrodes placed near the occipital area are ideally suited for capturing alpha-wave activity associated with relaxation and calm states.

\begin{figure}[htbp] 
    \centerline{
    \includegraphics[scale=0.6]{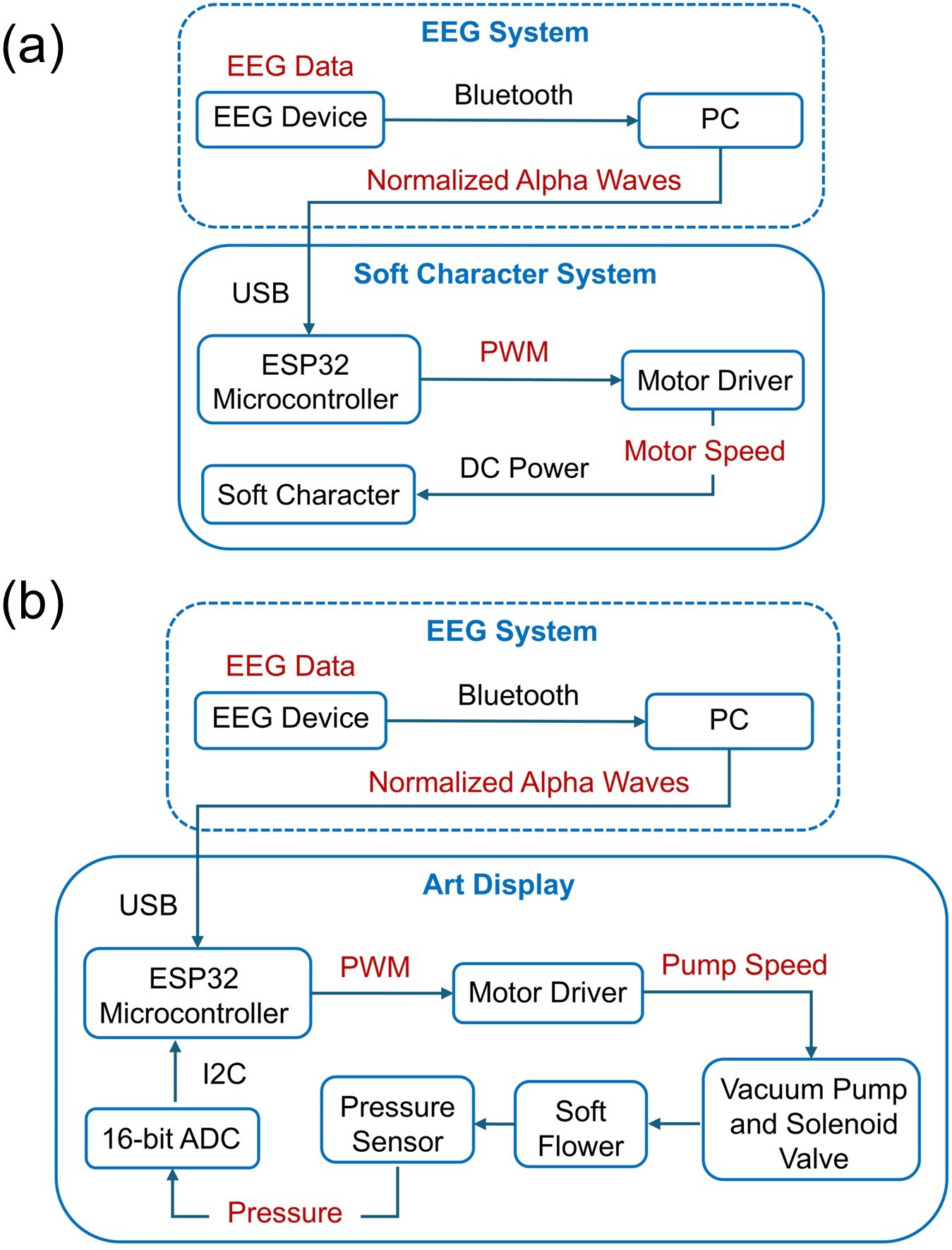}}
    \caption{(a) Soft character hardware system overview. (b) Hardware system overview of the soft flower as an art display.}
    \label{sys_overview_fig}
    \vspace*{-3mm}
\end{figure}

In this study, EEG data was acquired using the Unicorn Hybrid Black EEG device (\textit{g.tec medical engineering GMBH}) at a sampling rate of 250 Hz. To simplify data processing and reduce system complexity, data was collected from the electrode placed at the Oz position of the 10-20 system. The collected data was first processed through a 1-40 Hz bandpass filter to remove unwanted noise and retain only the key signal frequencies.

The system is developed using Python to facilitate real-time EEG data processing and analysis. The system analyzes the data using a fixed-length rolling window during real-time processing. The window length is 500 data points, and a 1/2 overlap is applied to enhance continuity. The Fast Fourier Transform (FFT) was applied in each window to convert the time-domain signals into the frequency domain, generating the Power Spectral Density (PSD). The analysis specifically focuses on the Alpha wave frequency range of 8-13 Hz, and the system monitors changes in power within this band.

To ensure that the detected signal corresponds to alpha waves, the system only considers the peak in the 6-20 Hz range, if it falls within the 8-13 Hz interval. Once a peak in the Alpha band exceeds the predefined threshold, the system records and responds to this activity. The PSD values of the user's Alpha waves ($A_{PSD}$) were normalized to a range of 0-100, and this value is sent to the microcontroller (Xiao ESP32-S3, Seeed Studio, Inc) via serial port at a baud rate of 115200.

This setup allows the system to monitor alpha wave activity in real-time, providing precise analysis and rapid feedback on the user's EEG signals.

\begin{figure*}[htbp]
    \centerline{
    \includegraphics[scale=0.51]{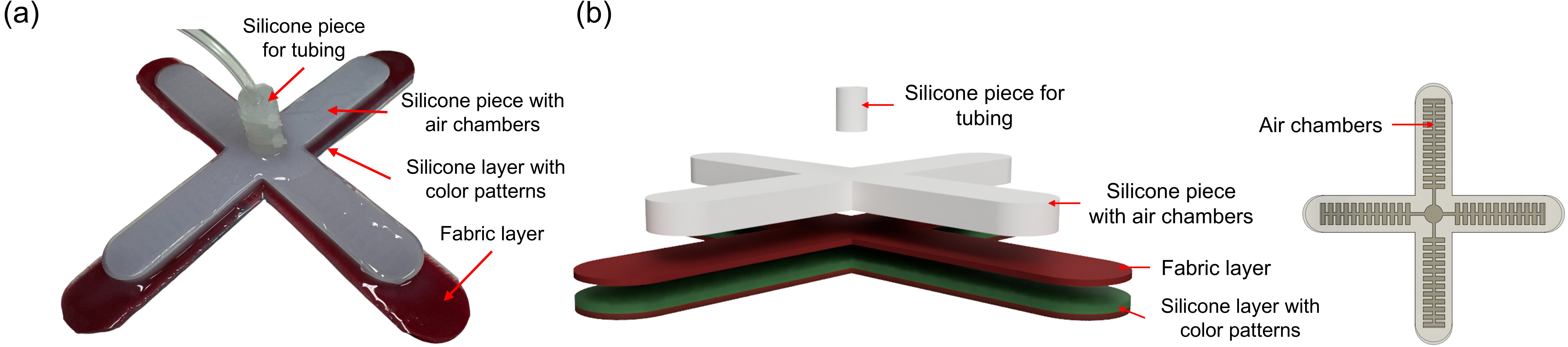}}
    \caption{(a) Fabricated soft flower with detailed labeling. (b) The exploded view of the soft flower with each layer labeled. The hollow internal air chambers are also shown here. }
    \label{fig:enter-label}
\end{figure*}

\subsection{Soft embodiment 1: a soft character}\label{soft_char_sec}

Emotions can be conveyed through several mediums. Arts and entertainment are among the most common ways people express their emotions. Soft characters intended for entertainment can be used to express emotions \cite{degiorgi2017puffy}. In this paper, the Wacky Waving Inflatable Tube Gal (by \textit{Gemma Correll and Conor Riordan}) was used as a soft character to convey emotions through a brain-mediated approach.  

The normalized alpha waves mentioned in section ~\ref{eeg_fft_sec} were used to do this. The PSD values of the alpha waves increase if the user is calm, which can be conveyed with a dancing motion. If the person is stressed, the soft character remains deflated. The frequency of the dancing is low for intermediate states. 

 The values of $A_{PSD}$ were mapped between 0-255, the resolution of the microcontroller's Digital to Analog Converter (DAC). These values were used to change the duty cycle of the Pulse Width Modulation (PWM) signal ($S_{PWM}$) sent to the L298N motor driver which controls the speed of the soft character's motor via its DC power socket. The normalized alpha waves mapped between 0 and 100 are the PWM signals' duty cycles.

\begin{equation}\label{pwm_sig_eq}
    S_{PWM} = \alpha A_{PSD}, 
\end{equation}

where $\alpha=2.55$ is a scaling factor that maps $A_{PSD}$ with  $S_{PWM}$. The speed of the motor is proportional to the dancing frequency of the soft character. Refer to Fig. 2(a) for an overview of the implemented system to control the soft character via EEG signals.

\begin{figure*}[htbp]
    \centerline{\includegraphics[scale=0.5]{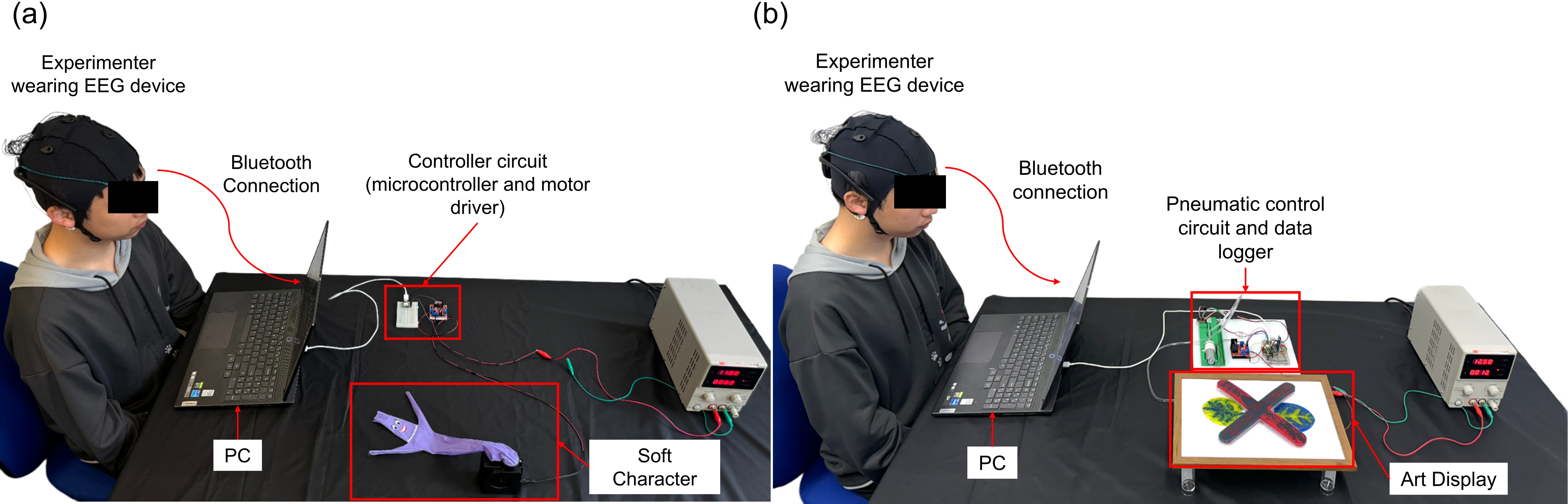}} 
    \caption{Experiment set-up (a) Soft character (b) Soft flower as an art display. }
    \label{fig:placeholder}
    \vspace*{-3mm}
\end{figure*}

\subsection{Soft embodiment 2: a soft flower in terms of an art display}

Artists use several styles of art to communicate their emotions to the audience, creating a powerful connection between their emotional state and those who view their work. Art displays employ unique shapes, colors, and objects to tell a story and convey emotions \cite{jin2024exploration}. In this paper, a human emotion-mediated art display was developed to communicate the user's emotions through interactive soft robots. The art display consists of a soft robotic gripper presented as a flower with leaves fabricated from silicone as additional details. 

The bright-colored soft flower is an interactive medium to show the user's emotional state. The maximum pressure applied to the soft flower, inflation, and deflation times, are dependent on the values of $A_{PSD}$. Closed-loop control was implemented with a PID algorithm to control and maintain the air pressure within the soft flower. The pressure setpoint for the soft flower ($y^{p}_{sp}$) was mapped with $A_{PSD}$ in the following way,

\begin{equation}
     y^{p}_{sp} = \beta A_{PSD},
\end{equation}

where $\beta=0.15$ is the scaling factor to map $A_{PSD}$ with $y^{p}_{sp}$. The value of $\beta$ was determined with the maximum pressure (135\textit{kPa}) when the soft flower is fully pressurized and the minimum pressure (120\textit{kPa}) when the soft flower is in the normal state. 

The inflation and deflation time varied based on the values of $A_{PSD}$. This enabled the creation of an interactive motion to display the user's emotional state. For example, when the person is calmer, the soft flower will inflate and deflate at a slower, prolonged rate. Conversely, when the person is stressed, the inflation and deflation cycle will speed up, reflecting the heightened tension. The inflation time ($t_{inflation}$) and deflation time ($t_{deflation}$) were mapped with $A_{PSD}$ in the following way,

\begin{equation}
    t_{inflation} = \gamma A_{PSD},
\end{equation}
\begin{equation}
    t_{deflation} = t_{inflation} + 0.5,
\end{equation}
where $\gamma=0.02$ is a scaling factor map $t_{inflation}$ with $A_{PSD}$. The value of $\gamma$ was determined with the maximum inflation time ($t^{max}_{inflation}=2.8s$) and minimum inflation time ($t^{min}_{inflation}=0.8s$). These values were selected from pilot trials. The solenoid valve is closed during the inflation cycle, and the solenoid valve is opened during the deflation cycle. 

The soft flower and leaves were fabricated with Ecoflex-0050 (Smooth-On, Inc.). The soft flower consists of three main layers. The first layer consists of a silicone piece that contains hollow cavities that act as air chambers during the inflation of the soft flower. The second layer is a cotton fabric that allows the soft flower to bend when inflated. The third layer is a thick silicone piece. The fabric was placed in-between the first and third layers and silicone was used to seal and bond all pieces together. Refer to Fig. 3 for a detailed overview of the fabricated soft flower. The soft flower was actuated by a small 12V vacuum pump (Jadeshay, Inc.) and solenoid valve (Source Mapping, Inc.), which an L298N motor driver controlled. 

Pressure sensors (MPXHZ6400A, Freescale Semiconductors, Inc.) were used to acquire the air pressure within the soft flower for closed-loop control. A 16-bit analog-to-digital converter (ADC) (ADS1115, DFRobotics, Inc.) with a timer interrupt was used to acquire pressure sensor signals at a sampling rate of 100\textit{Hz}. A moving average filter (window size of 10 samples) was applied to the pressure data to reduce noise. Refer to Fig. 2(b) for an overview of the hardware system of the soft flower as an art display.

\begin{figure*}[htbp] 
    \centerline{
    \includegraphics[scale=0.8]{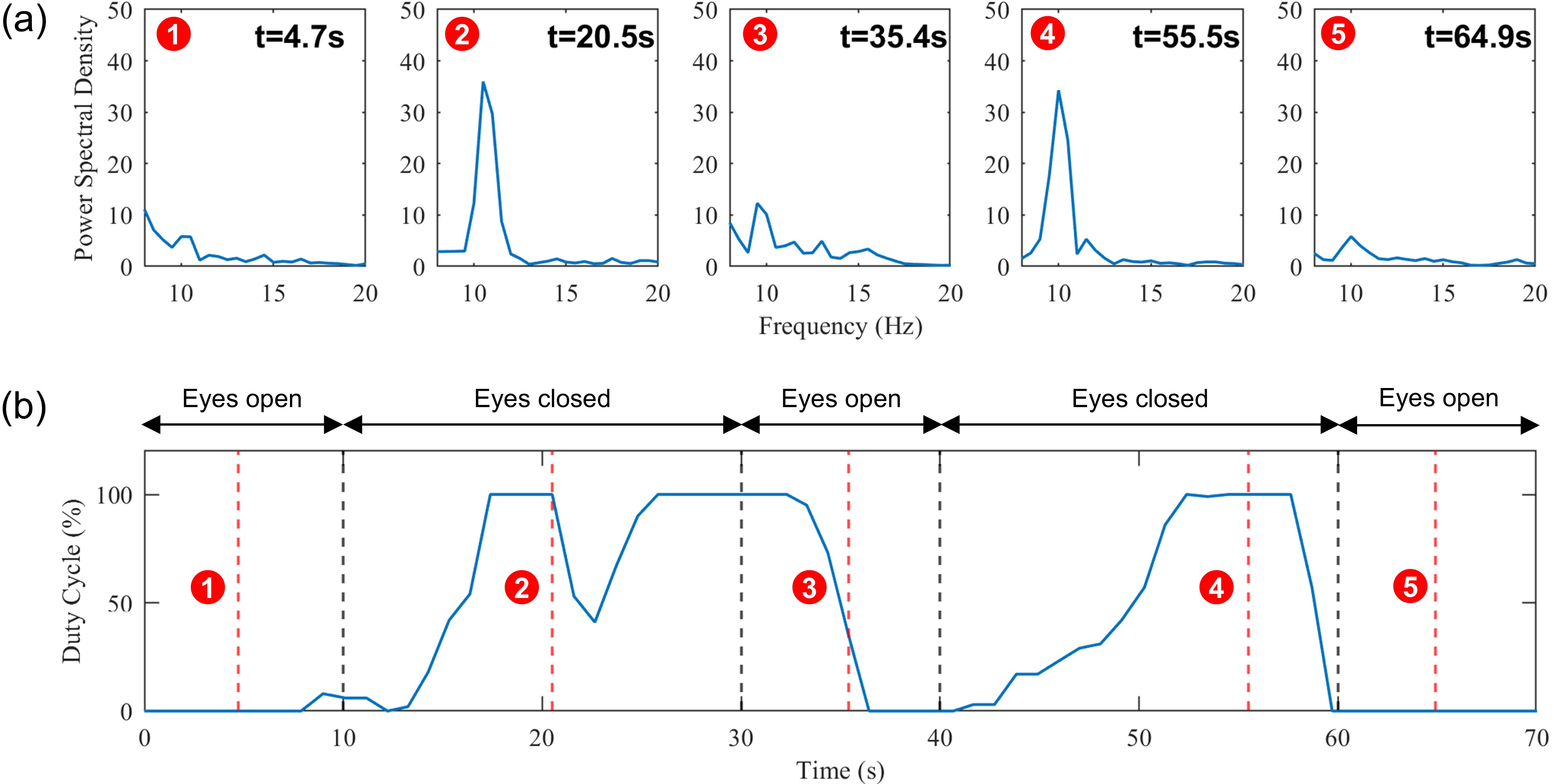}}
    \caption{The results are for the experiment conducted with the soft character. (a) Frequency vs PSD of the EEG alpha waves was selected at several instances when the system sent signals to the microcontroller (approximately every 1\textit{s}). (b) The variation in the duty cycle values sent from the PC throughout the experiment. The black dotted lines show when the experimenter was instructed to open and close his eyes. The numbers associate the alpha waves with the red dotted lines (instances for alpha waves in Fig. 5(a)) }
    \label{sys_overview_fig}
\end{figure*}
\begin{figure*}[htbp] 
    \centerline{
    \includegraphics[scale=0.8]{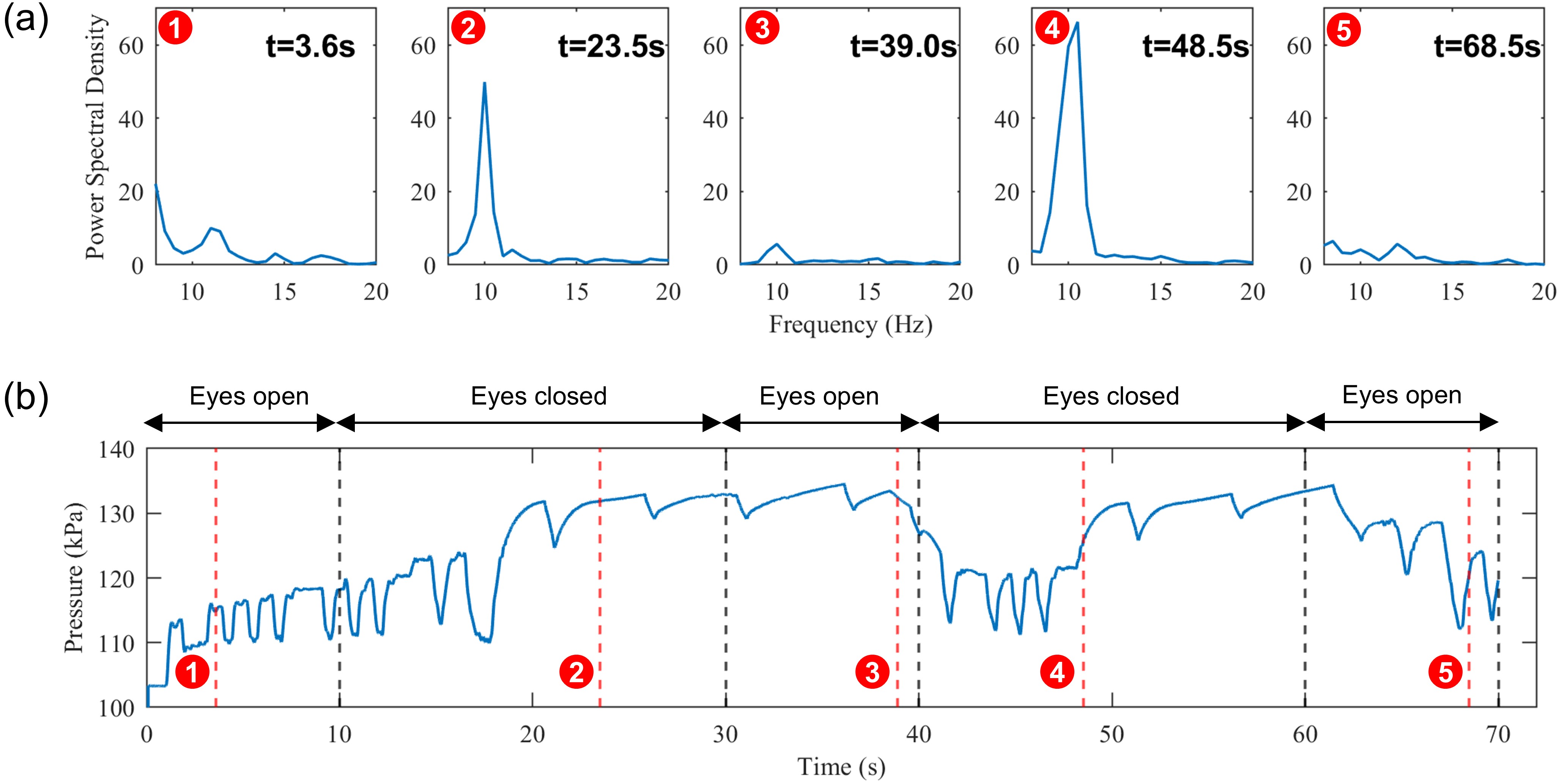}}
    \caption{The results are for the experiment conducted with the soft flower. (a) Frequency vs PSD of the EEG alpha waves was selected at several instances when the system sent signals to the microcontroller (approximately every 5\textit{s}). (b) The time vs pressure change in the soft flower during inflation and deflation. The black dotted lines show when the experimenter was instructed to open and close his eyes. The numbers associate the alpha waves with the red dotted lines (instances selected for alpha waves in Fig. 6(a).) }
    \label{sys_overview_fig}
    \vspace*{-3mm}
\end{figure*}

\section{EXPERIMENTS AND RESULTS }

% \textbf{Experiment setups for two soft embodiments (i.e., soft character and soft flower) are shown in Fig. X. The experimenter (22 years, male) wore the EEG system and sat on a chair. A buzzer was added to the pneumatic control circuit only for this experiment. The user was instructed to keep his eyes open at the start of the experiment.  He was asked to open his eyes at the beginning of the experiment. Then, at 10 seconds, the buzzer sounds indicating to close his eyes and try to remain calm. At 20 seconds, the buzzer sounds indicating to open his eyes again and wait for another 10 seconds. Therefore, the entire duration of the experiment was 30 seconds. %Timing instructions for closing and opening the eyes were provided using a metronome. 
% We repeated this procedure separately for each soft embodiment. During the experiments, we acquired raw EEG signals, PSD data, normalized alpha waves, output from the BCI (during soft character experiments), and pneumatic pressure data for further analysis.}

Experiment setups for two soft embodiments (i.e., soft character and soft flower) are shown in Fig. 4. The experimenter (22 years, male) wore the EEG device and sat on a chair. The experimenter was instructed to keep his eyes open at the start of the experiment. Then, the experimenter was instructed to close his eyes for 10 seconds and remain calm for 20 seconds. Again, the experimenter was instructed to open his eyes for another 10 seconds. Followed by 20 seconds of his eyes closed and 10 seconds of his eyes open. Therefore, the duration of the trial was 70 seconds. Timing instructions for the experimenter to close and open his eyes were provided verbally and timing was done with a stopwatch. We repeated this procedure separately for each soft embodiment. During the experiments, we acquired raw EEG signals, PSD data, normalized alpha waves, output from the BCI (during soft character experiments), and pneumatic pressure data for further analysis. The pressure data acquired during the experiment was stored within an SD card using a data logger (Micro SD card module).  

Fig. 5 shows the results during an experiment with the soft character. Fig. 5(a) illustrates how the EEG alpha signal varied during the experiments. Fig. 5(b) shows the duty cycle values sent from the PC to the microcontroller. We can see the corresponding duty cycle changes with the respective alpha waves. When the experimenter opened their eyes, the alpha waves were not obvious, the duty cycle was low, and the soft character's movements amplitude was small. As the user closed their eyes and relaxed, the alpha wave intensity increased rapidly, and the soft character's movement speed and amplitude increased significantly.

Fig. 6 shows the results during an experiment with the soft flower. Fig. 6(a) illustrates how the EEG alpha signal varied during the experiments. Fig. 6(b) shows the pressure change in the soft flower during the inflation and deflation cycles. Inflation times correspond to increasing pressure and deflation times correspond to decreasing pressure in Fig. 6(b). When the experimenter opened his eyes, the alpha waves were not obvious, the air pressure within the soft flower was small, and the inflation and deflation times were low. When the experimenter closed his eyes and relaxed, the alpha wave intensity increased rapidly. Inflation time, deflation time, and air pressure within the soft flower increased significantly. We observed from the results, that the pressure was below the minimum pressure value (120\textit{kPa}) when the experimenter's eyes were open. Reducing the deflation time or not opening the solenoid valve when the current pressure was below the minimum pressure would have prevented the pressure from falling below the minimum pressure. 
% This shows the $y^{p}_{sp}$ value increases with the strength of the alpha waves. Additionally, a correlation can be observed between, the strength of the alpha waves, $t_{inflation}$ and $t_{deflation}$.  

\subsection{Limitations and Future Work}

The main focus of this paper was to explore the idea of integrating human emotions with soft robotics and arts. Therefore, the methodologies were tested with one experimenter. In future research, this could expand to multiple participants to understand the relationship between diverse emotions, arts, and soft robotics. In the future, a qualitative study can be conducted to deepen others' understanding of human emotion-mediated soft robotic arts. Technical challenges such as latency in real-time signal processing and integration of multi-modal physiological signals or enhanced actuation mechanisms can be addressed in our future work.   

\section{CONCLUSIONS}

In this paper, we explored the conceptual idea of combining human emotions with soft robotics to create human emotion-mediated soft robotic arts. We have demonstrated how human emotion, via brain alpha waves in terms of EEG signals, can be mapped to dynamically control two soft embodiments: a soft character and a soft flower as an art display. We highlighted the inherent benefits of soft embodiments in the context of art. Through this integration, both the soft flower and the soft character showed the potential of soft robotics to bridge technology and artistic expression. This creates an innovative, responsive platform where human emotions, art, and robotics converge. By translating real-time human emotion via brain signals into the adaptive behaviors of soft robots, we envisage that this work opens a new frontier in nonverbal emotion-based communication, offering a unique platform that blends human experience, technology, and art. The work proposed in this paper can be implemented in many real-world applications such as potential use in public art galleries, interactive environments such as children's play areas, and entertainment events.

In future research on emotion-mediated soft robotics, several key directions deserve attention. First, improving the accuracy of emotion recognition is crucial. This can be achieved by improving emotion recognition techniques and integrating multiple physiological signals to help robots better perceive human emotions. Additionally, expanding the ways robots express emotions will be important. By adding dynamic elements, such as changes in color and shape, soft robots can convey emotions more precisely. At the same time, optimizing the signal processing speed will improve the real-time response speed of interaction, thereby ensuring the matching of human emotions and robot responses. Advancing these areas will support further advancement in human emotion-mediated soft robotic arts.

\addtolength{\textheight}{-12cm}   % This command serves to balance the column lengths

%%%%%%%%%%%%%%%%%%%%%%%%%%%%%%%%%%%%%%%%%%%%%%%%%%%%%%%%%%%%%%%%%%%%%%%%%%%%%%%%

%%%%%%%%%%%%%%%%%%%%%%%%%%%%%%%%%%%%%%%%%%%%%%%%%%%%%%%%%%%%%%%%%%%%%%%%%%%%%%%%

%%%%%%%%%%%%%%%%%%%%%%%%%%%%%%%%%%%%%%%%%%%%%%%%%%%%%%%%%%%%%%%%%%%%%%%%%%%%%%%%

\bibliographystyle{IEEEtran}  
\bibliography{IEEEexample}

\end{document}